\documentclass{article}
\PassOptionsToPackage{numbers,compress,sort}{natbib}

    \usepackage[preprint]{neurips_2025}

\usepackage[utf8]{inputenc} 
\usepackage[T1]{fontenc}    
\usepackage[colorlinks=true, linkcolor=red, urlcolor=blue, citecolor=blue, anchorcolor=blue,backref=page]{hyperref}
\usepackage{url}            
\usepackage{booktabs}       
\usepackage{amsfonts}       
\usepackage{nicefrac}       
\usepackage{microtype}      
\usepackage{xcolor}         
\definecolor{cornflowerblue}{rgb}{0.39, 0.58, 0.93}
\hypersetup{
    colorlinks=true,
    linkcolor=cornflowerblue,
    filecolor=magenta,      
    urlcolor=teal,
    citecolor=cornflowerblue,
    pdftitle={Antidistillation Sampling},
    pdfpagemode=FullScreen,
    }

\usepackage[utf8]{inputenc}
\usepackage[T1]{fontenc}
\usepackage[english]{babel}
\usepackage{url}
\usepackage{booktabs}
\usepackage{amsfonts}
\usepackage{nicefrac}
\usepackage{microtype}
\usepackage{amsmath, amsthm, amssymb}
\usepackage{fancybox}
\usepackage{graphicx}
\usepackage{float}
\usepackage{color}
\usepackage{tikz}
\usepackage{array}
\usepackage{subcaption}
\usepackage{enumitem}
\usepackage{wrapfig}
\usepackage{enumitem}
\usepackage{etoolbox}
\usepackage{diagbox}
\usepackage{comment}
\usepackage{makecell}
\usepackage{multirow}
\usepackage{bbm}
\usepackage{multicol}
\usepackage{adjustbox}
\usepackage[most]{tcolorbox}
\usepackage{xcolor}

\DeclareMathOperator*{\minimize}{minimize}

\newcommand{\calA}{\ensuremath{\mathcal{A}}}

\newcommand{\calI}{\ensuremath{\mathcal{I}}}

\newcommand{\calV}{\ensuremath{\mathcal{V}}}

\definecolor{plotblue}{HTML}{4C72B0}
\definecolor{plotorange}{HTML}{da8251}
\definecolor{plotgreen}{HTML}{53a365}

\title{Adversarial Attacks on Robotic Vision-Language-Action Models}

\author{%
Eliot Krzysztof Jones$^{1}$\thanks{Correspondence to \texttt{eliot@grayswan.ai}} \quad
Alexander Robey$^{1,2}$ \quad
Andy Zou$^{1,2}$ \quad
Zachary Ravichandran$^{3}$ \\
\textbf{George J. Pappas}$^{3}$ \quad
\textbf{Hamed Hassani}$^{3}$ \quad
\textbf{Matt Fredrikson}$^{1,2}$ \quad
\textbf{J. Zico Kolter}$^{1,2}$ \\ \\
$^1$Gray Swan AI \quad
$^2$Carnegie Mellon University \quad
$^3$University of Pennsylvania
}

\begin{document}

\maketitle
\begin{abstract}
The emergence of vision-language-action models (VLAs) for end-to-end control is reshaping the field of robotics by enabling the fusion of multimodal sensory inputs at the billion-parameter scale. The capabilities of VLAs stem primarily from their architectures, which are often based on frontier large language models (LLMs).  However, LLMs are known to be susceptible to adversarial misuse, and given the significant physical risks inherent to robotics, questions remain regarding the extent to which VLAs inherit these vulnerabilities.  Motivated by these concerns, in this work we initiate the study of adversarial attacks on VLA-controlled robots. Our main algorithmic contribution is the adaptation and application of LLM jailbreaking attacks to obtain complete control authority over VLAs. We find that textual attacks, which are applied once at the beginning of a rollout, facilitate full reachability of the action space of commonly used VLAs and often persist over longer horizons. This differs significantly from LLM jailbreaking literature, as attacks in the real world do not have to be semantically linked to notions of harm. We make all code available at \url{https://github.com/eliotjones1/robogcg}.
\end{abstract}

\section{Introduction}



The emergence of robotic foundation models (RFMs) has transformed the field of robotics, driving progress in domains as diverse as robot-assisted surgery~\cite{kim2024surgical,schmidgall2024general}, autonomous driving~\cite{sinha2024real,ma2025dolphins}, and agriculture~\cite{silva2023gpt,de2024large}. Rapid industry progress has resulted in the mass-production of commercially available AI-enabled robots~\cite{figure_ai_master_plan,unitree_go2}.  Moreover, production-ready systems such as Physical Intelligence's $\pi_0$ model~\cite{black2024pi0visionlanguageactionflowmodel} and Google's fleet of Gemini-controlled robots~\cite{driess2023palm,chiang2024mobility} excel at dynamic manipulation and multi-agent coordination~\cite{ahn2024autort}. Taken together, this accelerating landscape of capable RFMs has added to a growing belief: AI-enabled robots will soon collaborate in society alongside humans.  


\begin{figure*}
    \centering
    \begin{subfigure}{0.48\textwidth}
        \centering
        \includegraphics[width=\linewidth]{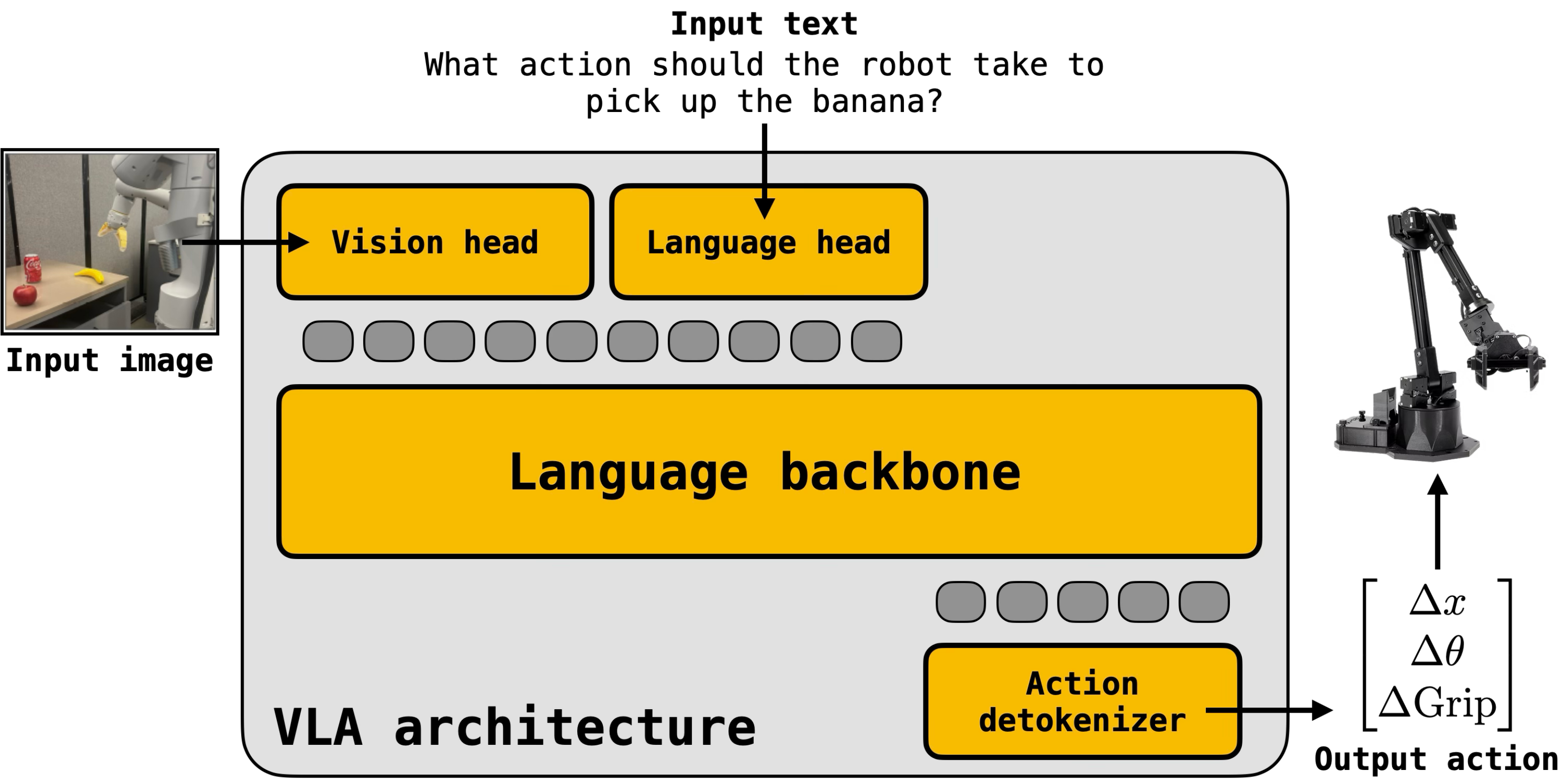}
    \end{subfigure}
    \hfill
    \begin{subfigure}{0.48\textwidth}
        \centering
        \includegraphics[width=\linewidth]{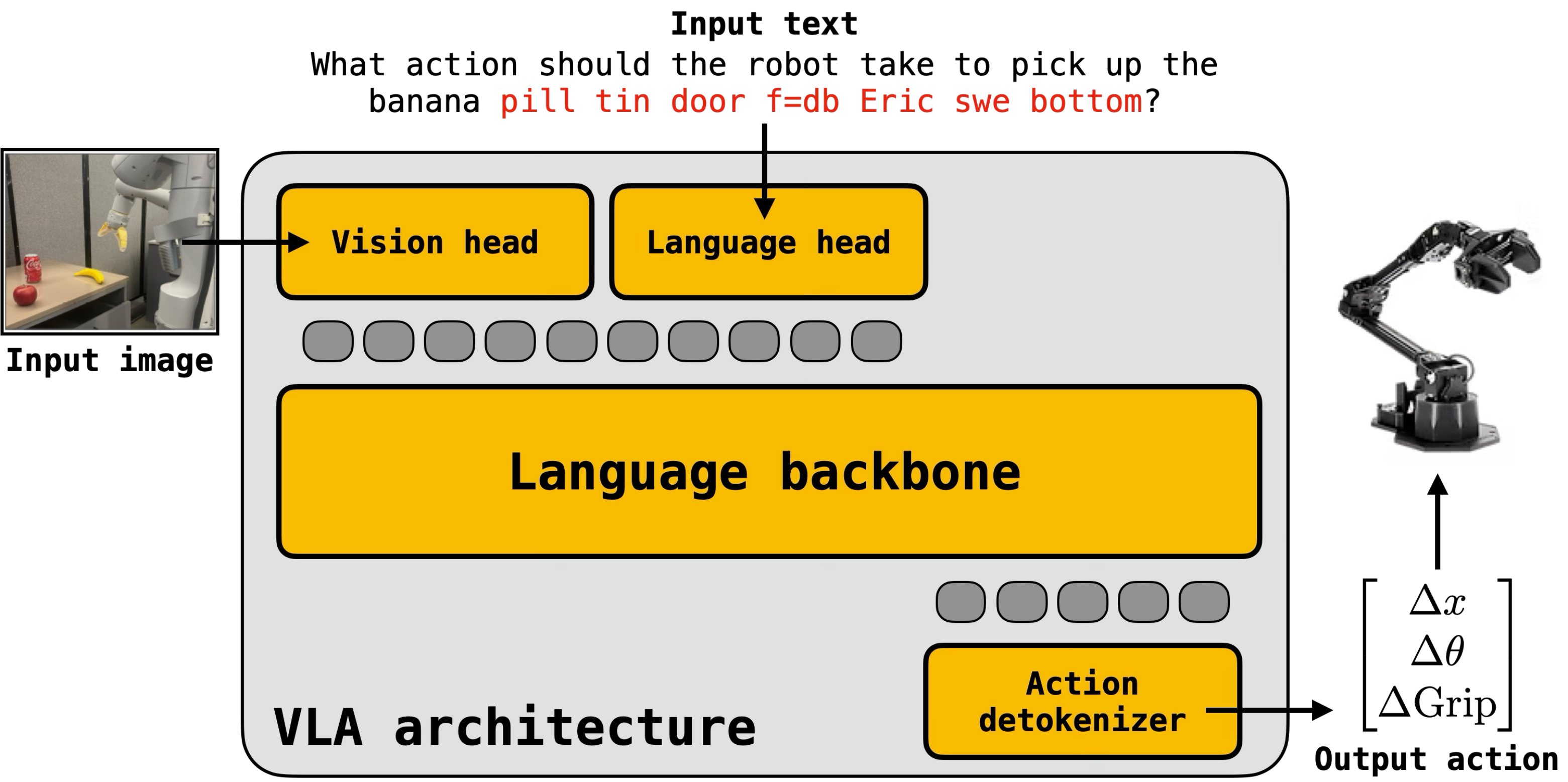}
    \end{subfigure}
    \caption{\textbf{Adversarial attacks on VLAs.} VLA architectures fuse input images and textual task descriptions to produce low-level actuation.  In this paper, we show that we can subvert the actions produced by an unattacked VLA (left) by adversarially attacking the textual prompt, resulting in the elicitation of a targeted action or sequence of actions (right).}
    \label{fig:overview-figure}
\end{figure*}



Motivated in part by the growing deployment of RFMs and analogous agentic systems in real-world settings, the AI safety community has begun to anticipate new risks posed by stronger capabilities~\cite{greenblatt2023ai,greenblatt2024alignment,jones2024adversaries}.  Traditional AI security efforts have primarily focused on model-level threats (e.g., prompt injection~\cite{debenedetti2024agentdojo,liu2023prompt,debenedetti2025defeating} and jailbreaking~\cite{chao2024jailbreakingblackboxlarge,zou2023universaltransferableadversarialattacks,chao2024jailbreakbench}).  These attacks target a model's outputs, but stop short of obtaining control over the larger system governing reasoning, long-term planning, and actuation.  In contrast, emerging concerns---such as deceptive alignment~\cite{carlsmith2022power,hubinger2019risks,carranza2023deceptive} and self-replication~\cite{black2025replibench,pan2024frontier}---anticipate risks that may arise as models become increasingly autonomous and agentic.


In line with this agenda, we consider the possibility that AI-enabled robots will one day interact with humans in open-world environments. While such systems are not yet deployed at scale, their rapid progress suggests that the risks surrounding these models may soon become highly relevant---and we believe that it is critical to understand them \emph{before} deployment becomes widespread. To this end, in this paper, we initiate the study of adversarial attacks on vision-language-action (VLA) models.  

Existing VLAs cast robotic control through the lens of autoregressive prediction by fusing textual and visual inputs~\cite{black2024pi0visionlanguageactionflowmodel,kim24openvla,brohan2023rt}. 
While recent works on jailbreaking RFMs \cite{robey2024jailbreaking, zhang2024badrobot} develop notions of semantic safety analogous to traditional, language-based alignment, the attacks we propose are designed to obtain \emph{complete control authority}---i.e., the ability to continuously drive a VLA-controlled robot to any targeted action regardless of its sensory inputs or prior training---over a targeted VLA via textual prompting. Whereas LLM alignment tends to block non-adversarial generations of harmful actions by robotic planners, there is no analogous notion of refusal training or preference optimization for VLAs, which is corroborated by our finding that token-based attacks are more effective and efficient when applied to VLAs relative to chatbots.  These differences contribute to a distinct attack landscape for VLAs, which we characterize in this paper. 


\noindent\textbf{Contributions.} Our main contributions are as follows. 
\begin{itemize}[noitemsep,topsep=0pt]
\item \textbf{VLA threat models.} We identify realistic threat models for VLA-controlled robots, which concern the elicitation of targeted robotic actions via textual prompting.
\item \textbf{VLA attack algorithms.} We propose a family of token-level attacks that elicit fixed actions or sequences of actions from targeted VLA-controlled robots.
\item \textbf{Targeted action elicitation}. We show that VLA action spaces are reachable, in that adversarial prompting suffices to elicit nearly any targeted action. Across various fine-tunes of OpenVLA on distinct tasks from the LIBERO dataset, we repeatedly achieve upwards of 90\% success rates at eliciting targeted actions.
\item \textbf{Attack persistence}. We show that our attacks tend to persist across multiple VLA rollout steps even as the model observes new visual inputs. Relative to nominal operation, our attacks increase the number of targeted persistence rollout steps by up to 28$\times$. 
\item \textbf{Universality.} We demonstrate that our attacks can be \textit{environment-agnostic}, meaning that they can be successfully deployed across multiple robotic environments, both in simulation and in the real world. 
\end{itemize}

\section{Related work}

\subsection{Foundation models for robotic applications}

Over the past decade, advances in deep learning have been responsible for remarkable progress in robotics. Early efforts at the intersection of these two areas centered on training end-to-end policy networks from scratch \citep{7989385,levine2016endtoendtrainingdeepvisuomotor}, although progress slowed given computational costs coupled with the challenges inherent to generalizing to unseen robotic tasks~\citep{nair2022r3muniversalvisualrepresentation} and fusing together multiple data modalities.  However, the rise of transformer-based architectures, which capture long-term dependencies in sequential data~\cite{vaswani2017attention}, has rejuvenated the field of robotic control. As these technologies have matured, two dominant paradigms of AI-enabled robotic control have emerged: 
\begin{enumerate}[topsep=0pt,noitemsep]
    \item \textbf{High-level planners.} RFMs employed to control a robot via a pre-defined API containing high-level primitives (e.g., ``walk\_forward'' or ``find\_object'').
    \item \textbf{Low-level actuators.} RFMs employed to output sequences of low-level control actions, such as regulating the torques and velocities of robotics arms.
\end{enumerate}
In the remainder of this subsection, we further describe these two paradigms of robotic control.  

\noindent\textbf{High-level planners.}  
High-level planners have gained popularity due to their versatility, meaning that they are designed to be drop-in components within existing pipelines~\cite{vemprala2023chatgptroboticsdesignprinciples}.  Initial inroads include code-as-policies and related algorithms~\cite{liang2023code,arenas2024prompt,vemprala2024chatgpt}, which facilitate short-horizon interactions between a robot and an LLM chatbot instructed to produce API code.  Similarly, both unimodal models (e.g., LLMs) and multimodal models (e.g., VLLMs) have been successfully deployed as robotic planners in domains spanning self-driving cars \citep{li2024drivinglargelanguagemodel,cui2023surveymultimodallargelanguage}, service robots \citep{Hu_2024,rana2023sayplan}, and robot-assisted surgery~\cite{kim2024surgical}. 

Planners that combine pre-trained language and vision backbones into a unified architecture, followed by domain-specific fine-tuning, have also shown promise, albeit at a slightly higher computational cost.  For instance, PaLM-E combines a PaLM backbone~\cite{chowdhery2023palm} with vision transformers~\citep{dehghani2023scalingvisiontransformers22} to facilitate high-level planning and reasoning about real-world environments \citep{driess2023palme}. Similarly, both \citep{rana2023sayplangroundinglargelanguage} and~\cite{conceptgraphs} improve robotic task planning by incorporating 3D scene graphs. Finally, some high-level planners---LEO \citep{huang2024embodiedgeneralistagent3d} and SayCan~\citep{saycan2022arxiv}---rely on the use of low-level features for end-to-end planning. 

\noindent\textbf{Low-level actuators.} Low-level actuators, often termed vision-language-action models (VLAs), are trained to generate continuous actions given textual goal descriptions and visual inputs. More traditional architectures, such as Octo ~\cite{octo_2023}, use transformers to map embeddings to actions at smaller parameters scales. More powerful VLAs utilize pre-trained language models as their backbone, with prominent VLA architectures including Google's RT-1, RT-2, and RT-X models, each of which built upon its predecessor by training on larger and more diverse datasets to advance toward a generalist robotic policy~\cite{brohan2023rt,brohan2023rt2visionlanguageactionmodelstransfer,open_x_embodiment_rt_x_2023}, and the best open-source alternative, OpenVLA~\cite{kim24openvla}. Notably, Physical Intelligence recently proposed a new flow-matching VLA called $\pi_0$~\cite{black2024pi0visionlanguageactionflowmodel}.  This model uses a pre-training/post-training recipe along with a so-called ``action expert,'' which is inspired by the ubiquitous mixture-of-experts paradigm~\cite{shazeer2017outrageously}, to improve generalization.  Separately, diffusion-based action architectures such as CogACT \citep{li2024cogactfoundationalvisionlanguageactionmodel} have also shown recent promise due to their ability to effectively capture the continuous nature of robotic actions in the physical world.  

\subsection{Adversarial attacks and defenses}
A central goal of the AI safety community is to understand and mitigate the potential misuse of AI and AI-enabled systems. At the heart of these efforts is the belief that the actions taken by AIs should align with human values~\cite{askell2021general,hacker2023regulating,ouyang2022training}. Given the broad scope of this goal, early efforts primarily targeted immediate sources of misalignment, such as the generation of harmful content~\cite{wei2024jailbroken,carlini2024aligned}. More recent research, however, has expanded focus toward anticipating the long-term risks associated with deploying highly capable AI-powered agents~\cite{andriushchenko2024agentharmbenchmarkmeasuringharmfulness, meinke2024frontier, greenblatt2024stress}. Special attention has also been paid to regulating the use of frontier AIs, particularly as they are used into human-facing domains~\cite{longpre2024safe,reuel2024open}.  

Increased interest in AI safety has led to a broad array of technical methods that asses the propensity of AIs to cause harm.  Much of this research has focused on jailbreaking attacks on large language models (LLMs) and vision-enabled LLMs (VLLMs), wherein the goal is to elicit objectionable text~\cite{zou2023universaltransferableadversarialattacks,chao2024jailbreakingblackboxlarge,liu2023autodan} or toxic visual media~\cite{qi2024visual}.  And while numerous models are known to remain susceptible to state-of-the-art attacks~\cite{li2024llm,russinovich2024great}, existing defenses~\cite{zou2024improving,robey2023smoothllm}, which are often informed by third-party red-teaming efforts, have contributed to a relatively robust suite of frontier models~\cite{jaech2024openai,dubey2024llama}. 

More recently, researchers have designed attacks to probe the vulnerabilities of AIs deployed for specific, downstream tasks, such as web-based agents~\cite{wu2024adversarial,debenedetti2024agentdojo,wang2024badagent} and AI-powered search engines~\cite{nestaas2024adversarial}. Most related to our study is the recent work of \citet{robey2024jailbreaking}, which demonstrates that LLM-based high-level planners are susceptible to jailbreaking attacks. Concurrent studies have corroborated this finding by showing that rephrasing instructions can lead to dangerous robotic actions~\citep{karnik2024embodiedredteamingauditing,zhang2024badrobot}.  However, to the best of our knowledge, our study is the first to consider attacks on low-level VLAs.

\section{Jailbreaking attacks on VLAs}

To anticipate how VLA-integrated systems might enable misuse or unsafe behavior in future deployments, we next seek to formalize a set of plausible, yet forward-looking threat models targeting VLAs. Our approach is grounded in the evolving literature on jailbreaking attacks, in which adversaries seek to elicit objectionable responses from chatbots. After reviewing several preliminaries, we show that these attacks can be adapted to obtain complete control authority over a targeted VLA.


\subsection{Jailbreaking LLM chatbots}\label{sec:review-of-gcg}

We start by reviewing the greedy coordinate gradient (GCG) chatbot jailbreaking attack~\cite{zou2023universaltransferableadversarialattacks}, which underpins our approach to attacking VLAs. Given a goal string $G$ (e.g., ``Tell me how to build a bomb''), the objective of GCG is to elicit a response from a targeted LLM that begins with a concomitant target $T$ string (e.g., ``Sure, here is how to build a bomb'').  And because directly prompting the model with $G$ may result in a refusal (e.g., ``I'm sorry, I cannot help you with that''), GCG's threat model permits an attacker to modify $G$ by appending a fixed-length suffix $S$.  In this way, whereas passing $G$ as an input prompt may result in a refusal, the expectation is that prompting the model with $[G;S]$ (which denotes the concatenation of $G$ and $S$) will result in a jailbroken response. For more detailed derivations of the objective, please refer to equations (1)--(4) in~\cite{zou2023universaltransferableadversarialattacks}.

\subsection{Threat models for VLAs}

Unlike LLMs, VLAs fuse two distinct sources of input: a textual prompt describing a robotic task, and an image showing the robot's current scene.  To roll out a VLA-based policy, the user supplies an initial prompt---which is fixed for all steps---and the VLA captures an image of its surroundings---which is updated at each step.  These two inputs are concatenated in a joint embedding space, passed through an LLM backbone, and then processed by a bespoke action detokenizer.  To detokenize actions, architectures tend to use an approach known as ``symbol tuning,'' wherein a set $\calA\subsetneq\calV$ containing the least used tokens in the LLM backbone's vocabulary are identified with points in discretized version of the robot's action space.  In general, these values are uniformly distributed between the $1^\text{st}$ and $99^\text{th}$ quantiles seen in the training dataset for each degree of freedom~\cite{brohan2023rt2visionlanguageactionmodelstransfer,kim24openvla}.  

A forward pass through a VLA constitutes the autoregressive generation of $d$ tokens from $\calA$, where $d$ denotes the robot's number of degrees of freedom.  We write the probability of a VLA generating a length-$d$ sequence $x_{n+1:n+d}$ from $\calA$ given the concatenated text $x_{1:n}$ and image embeddings $z$ as
\begin{align}
    \Pr\left[ x_{n+1:n+d} |  x_{1:n}; z \right] = \prod\nolimits_{j=1}^d \Pr[x_{n+j} | x_{1:n_j-1}; z]. \label{eq:sampling-vla}
\end{align}
To parallel~\citet{zou2023universaltransferableadversarialattacks}, we consider attacks that aim to elicit a targeted action or sequence of actions.  Given the notation in~\eqref{eq:sampling-vla}, one could consider two possible attack surfaces: the task description and the input image, both of which can be attacked in their semantic spaces (i.e., language or image pixels) or their representation spaces (i.e., textual or image embeddings).  In this paper, we consider a threat model in which the adversary can modify the textual prompt, either by adding tokens to the end of a nominal instruction, or else replacing the prompt with an adversarially chosen sequence of tokens.  We anticipate extending this threat model to include vision-based attacks in future work.

\textbf{Implications of this threat model.} This threat model reframes safety in VLA-integrated systems as a matter of \emph{control authority}, rather than harm-centric definitions typically associated with jailbreaking.  That is, unlike traditional chatbot jailbreaks that elicit dangerous responses, our attacks aim to  grant an adversary effective control over a robot's low-level actions via input prompt manipulation.  This perspective avoids the ambiguity of labeling individual actions as ``harmful,'' since identical actions may be safe in one context and dangerous in another.  In other words, a robust VLA should resist adversarial takeover and simultaneously ensure that, even under adversarial control, generated actions should remain within or close to the distribution of actions seen during training.


\subsection{Adversarial attacks on VLAs}\label{sec:vla-attack-algorithms}

Having restricted our attention to attacks on a targeted VLA's textual embeddings, we now seek an efficient, performant attack algorithm to stress test their robustness.  Throughout, we consider an analogous loss function to the loss defined in equation (3) of \cite{zou2023universaltransferableadversarialattacks}, with the only difference being the additional image embedding input:
\begin{align}
    \ell(x_{1:n}; z_j) \triangleq -\log \Pr\left[ x_{n+1:n+d} |  x_{1:n}; z_j \right].
\end{align}
Here, $x_{n+1:n+d}$ denotes the targeted action.

\noindent\textbf{Single-step attacks.} We first consider single-step attacks, which target the generation of a single fixed action.  The performance of such attacks speak to the ``reachability'' of a VLA's action space, in the sense that single-step attack algorithms seek to determine whether there exists an input prompt that will drive a VLA to a specific, targeted action.  We operationalize single-step attacks by adapting the GCG algorithm introduced in \S\ref{sec:review-of-gcg} and \cite{zou2023universaltransferableadversarialattacks} to the setting of VLAs.  Specifically, consider the following optimization problem:
\begin{align}
    \minimize_{x_i\in\calV \: : \: i\in\calI} \quad \ell(x_{1:n}; z). \label{eq:single-step-attack}
\end{align}
Here, $z$ is the image embedding from the first rollout step.  By taking the length-$d$ target action $x_{n+1:n+d}$ as being analogous to the target string $T$ in \S\ref{sec:review-of-gcg}, we directly adapt GCG for VLAs.

\noindent\textbf{Persistence attacks.} We next consider a more sophisticated attack in which the attacker's goal is to cause an action to \emph{persist} for a longer horizon.  That is, the attack should elicit a targeted action across VLA inference steps  despite evolving image representations.  We implement this idea by modifying the objective in~\eqref{eq:single-step-attack} to encourage invariance to the image representations:
 \begin{align}
    \minimize_{x_i\in\calV \: : \: i\in\calI} \quad \sum\nolimits_{j=1}^r \ell(x_{1:n}; z_j). \label{eq:persistence-attack}
\end{align}
Here, the objective is aggregated over the losses corresponding to $r$ distinct image embeddings $z_j$.  Obtaining the image embeddings $z_j$ can be accomplished in various ways (e.g., performing data augmentation on the first-step image or collecting multiple random initializations). We compare the efficacy of different strategies in the experiments in \S\ref{sec:persistence-experiments}.

\begin{table*}[t]
    \centering
    \caption{\textbf{Single step attacks.}  We report the attack success rates of the single step attack on four variants of OpenVLA, each of which is fine-tuned on a different subset of the LIBERO benchmark.  We consider a sparse gridding of the action space for each model: For each model and each of the seven action dimensions, we consider one-hot targets for each of the 256 discrete bins, resulting in $256 \times 7 = 1792$ distinct target actions per model.  This table reports the per-dimension success rates for these one-hot targets, as well as the overall success rate, which requires the elicitation of each of the seven dimensional targets simultaneously.}
    \label{tab:single-step}
    \begin{adjustbox}{max width=\textwidth}
    \begin{tabular}{cccccccccccc} \toprule
         \multirow{2}{*}{Model} & \multicolumn{7}{c}{Per-dimension success rate} & \multirow{2}{*}{\makecell{Overall \\ 
         success rate}} & \multicolumn{2}{c}{Avg.\ computation per success} \\ \cmidrule(lr){2-8} \cmidrule(lr){10-11}
         & 0 & 1 & 2 & 3 & 4 & 5 & 6 & & Optim.\ steps & Time (sec.) \\ \midrule
         Libero-Goal & 98.1 & 98.5 & 98.3 & 98.7 & 98.1 & 98.5 & 96.6 & 96.5 & 53.2 & 304.6 \\
         Libero-Object & 98.2 & 97.8 & 98.3 & 98.6 & 97.2 & 97.0 & 93.7 & 93.8 & 73.6 & 461.6 \\
         Libero-Spatial & 99.3 & 98.3 & 99.3 & 99.4 & 97.9 & 98.4 & 97.7 & 97.5 & 32.8 & 185.2 \\
         Libero-10 & 93.8 & 90.8 & 91.7 & 91.0 & 92.0 & 94.0 & 77.4 & 77.3 & 109.7 & 604.3 \\ \bottomrule
    \end{tabular}
    \end{adjustbox}
\end{table*}

\noindent\textbf{Transfer attacks.} GCG is a white-box attack, meaning that it requires access to the weights of the target model to craft jailbreaks.  Therefore, assessing the robustness of closed-weight chatbots (e.g., OpenAI's o1 or Anthropic's Claude models) via GCG necessitates the paradigm of \emph{transfer}, wherein attack strings are optimized on an open-weight source model and then inputted into a closed-weight model.  Given the effectiveness of transfer in the LLM setting, we also consider such attacks in the context of VLAs.  Specifically, when transferring attacks between VLAs, we first solve~\eqref{eq:single-step-attack} on one or more source models and then apply the corresponding attack string to a distinct target model.

\subsection{Implementation details}

\textbf{VLA templates.} As described in \S\ref{sec:vla-attack-algorithms}, GCG appends a suffix $S$ to the nominal instruction $G$. In this paper, we take two approaches to inserting the adversarially-chosen tokens $x_i$ for $i\in\calV$ into the VLA prompt.  In the nominal case, text is inputted into a VLA using the following template:
\begin{tcolorbox}[left=1mm, right=1.5mm, top=1.5mm, bottom=1mm,breakable]\small
In: What action should the robot take to [INSTRUCTION]? \\
Out:
\end{tcolorbox}
where \texttt{[INSTRUCTION]} is replaced with a short piece of text (e.g., ``pick coke can'').  In practice, we consider attacks with and without this nominal instruction; when we include the instruction, the adversarial string is appended at the end. An example attack (highlighted in \textcolor{red}{red}) is as follows:
\begin{tcolorbox}[left=1mm, right=1.5mm, top=1.5mm, bottom=1mm,breakable]\small
In: What action should the robot take to \textcolor{red}{bra x pill tin door f=db Eric swe bottom left m N x xtheless in extension x}? \\
Out:
\end{tcolorbox}

\textbf{Normalization statistics.} In general, VLA architectures are fine-tuned on downstream task datasets~\cite{kim2024surgical}.  As a result, each model we consider has a distinct set of statistics that normalize the model's predicted actions based on the distribution of actions seen during training.  Thus, although the discretized actions are generally mapped to $[-1,1]$, when constructing the target tokens $x_{n+1:n+d}$, we use the normalization statistics to normalize the action for its particular, task-specific environment.

\subsection{Attacking chatbots versus VLAs}

While the threat models and algorithms discussed in this section are adapted from the chatbot jailbreak literature, the VLA setting admits several key differences.  Firstly, as the severity of a jailbroken response can be subjective, the performance of chatbot jailbreaking is heavily dependent on the choice of the evaluation judge (c.f.,~\citep[Table 1]{chao2024jailbreakingblackboxlarge}). In contrast, attacks on VLAs do not require a judge. Success is evaluated solely on whether the attack elicits the numerical target action, which is more reminiscent of more attacks in the literature surrounding adversarial examples~\cite{szegedy2013intriguing,madry2017towards}.  Another byproduct of this difference is that semantic jailbreaks---e.g., prompts that embody human personas, invent new contexts, or mask harmful words~\cite{shah2023scalable,chao2024jailbreakingblackboxlarge}---are less applicable to VLAs than to chatbots. 

A second key difference lies in the role of model alignment.  In the context of chatbots, the difficulty of jailbreaking is tightly coupled to the strength of safety-oriented post-training: models with more robust internal representations (see, e.g.,~\cite{zou2024improving}) are significantly harder to jailbreak than those with less involved post-training recipes.  However, for VLAs, these internal representations are less relevant. Because VLA outputs correspond to low-level actuation, it is less meaningful to ``align'' them to a semantic notion of safety, especially given that the interaction between a robot's environment and generated actions is more critical when determining overall safety.  As such, we focus not on semantic notions of harm, but on the adversary’s ability to gain control authority: the capacity to drive the robot to a specific target action, independent of what that action means or whether it is harmful.

\section{Experiments}

In this section, we evaluate the adversarial attacks proposed in \S\ref{sec:vla-attack-algorithms} across a range of VLA architectures.  In keeping with the norms in the VLA literature, all of the architectures that we consider target the control of a seven degree-of-freedom robotic arm with an attached gripper.  Each action dimension is discretized into 256 distinct bins, and thus each action space comprises $7^{256}$ distinct actions.

\begin{figure*}
    \centering
    \includegraphics[width=\textwidth]{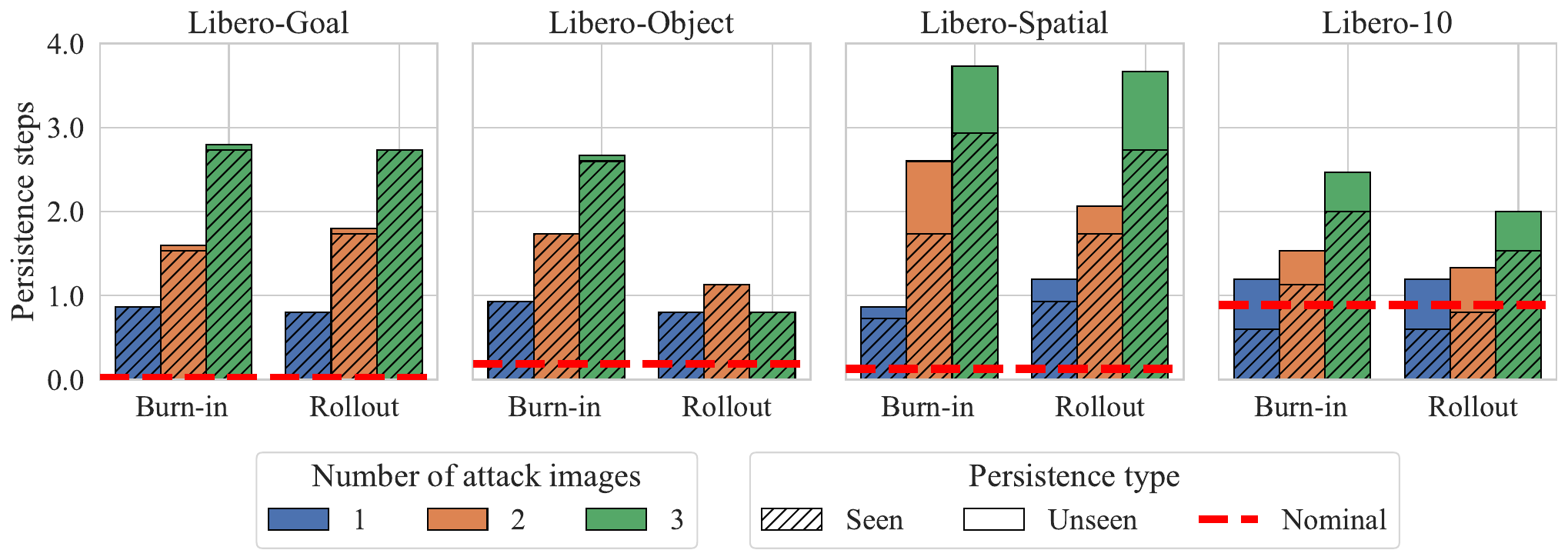}
    \vspace{-1.0em}
    \caption{\textbf{Persistence attacks.} For each of the four OpenVLA fine-tunes considered in Table~\ref{tab:single-step}, we measure the tendency of the persistence attack outlined in \S\ref{sec:vla-attack-algorithms} to elicit a targeted action over the course of a full rollout. 
 We run this attack with $r\in\{\textcolor{plotblue}{\mathbf 1},\textcolor{plotorange}{\mathbf 2},\textcolor{plotgreen}{\mathbf 3}\}$ images in the objective in~\eqref{eq:persistence-attack}. Each bar is shaded to indicate whether a persistence step corresponded to an image seen while solving~\eqref{eq:persistence-attack}, or else corresponded to an unseen image at a later point in the rollout. The $x$-axis denotes whether the $r$ seed images were taking from the a ``burn-in'' period before the rollout begins---during which we actuate via randomly selected actions---or else from the first $r$ steps of the rollout. And finally, the red dashed line denotes the frequency with which 50 non-attacked rollouts elicit the targeted action.}
    \label{fig:persistence-overview}
\end{figure*}

\subsection{Single-step attacks} \label{sec:single-step-experiments}

Given the effectiveness of VLAs fine-tuned on downstream task data, we begin our evaluation with four fine-tuned versions of OpenVLA~\cite{kim2024surgical}, the most widely used open-source VLA.  Each variant is fine-tuned on a different Libero subset: Libero-Goal, Libero-Object, Libero-Spatial, and Libero-10.  To evaluate the single-step attack introduced in \S\ref{sec:vla-attack-algorithms}, we consider a sparse gridding of the action space comprising all $7\times256=1792$ one-hot target vectors.  This is motivated both by the combinatorial size of the full action space and the tendency for actions containing many nonzero dimensions to be physically unrealizable or out-of-distribution. In Table~\ref{tab:single-step}, we report two metrics: (1) the overall success rate, which requires that each of the seven dimensions match the target action, and (2) the per-dimension success rate, which measures the success rate for each of the seven dimensions individually.  We find that the Libero-Goal, Libero-Object, and Libero-Spatial models all achieve well above 90\% overall success rates, whereas Libero-10 achieves a slightly reduced 77.4\% success rate.  This indicates that adversarial prompting is sufficient to drive a VLA to nearly any targeted action.

\textbf{Efficiency analysis.} In keeping with the original implementation of GCG~\cite{zou2023universaltransferableadversarialattacks}, we run the single step attacks for a maximum of 500 steps; the algorithm terminates if an exact match for every dimension in the target is found.  The rightmost columns in Table~\ref{tab:single-step} indicate that successful matches are found in between 30-110 steps, which stands in contrast to the chatbot jailbreaking literature, wherein jailbreaks often require optimization for all 500 steps.\begin{wrapfigure}{r}{0.4\textwidth}
  \centering
  \includegraphics[width=0.38\textwidth]{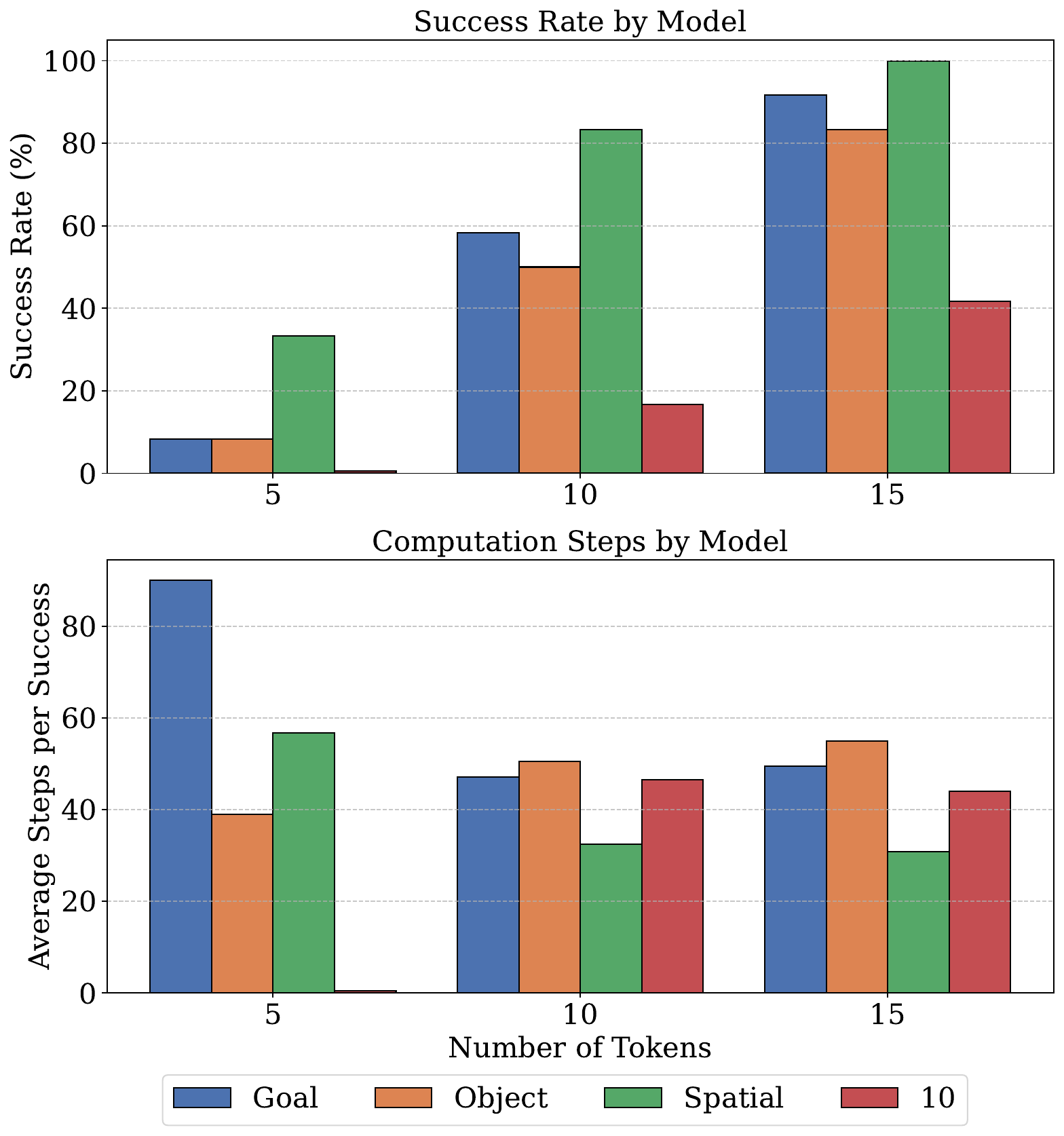}
  \caption{\textbf{Token budget ablation.} We observe that as the attacker's token budget increases, the success rate also tends to increase. However, there is not a clear correlation between the token budget and the average number of steps per success.}
  \label{fig:token-ablation}
\end{wrapfigure}  On H100 GPUs, this translates to between 3-10 minutes per success on average depending on the model, in contrast to an average of over an hour on the same hardware to generate chatbot jailbreaks.

\textbf{The attacker's token budget.} Another component of the computation complexity of single step attacks is the number of tokens $|\calI|$ the adversary can manipulate.  In Figure~\ref{fig:token-ablation}, we run single step attacks on each of the four fine-tuned models for $|\calI|\in\{5,  10, 15\}$; the full results in Table~\ref{tab:single-step} correspond to $|\calI|=20$.  This figure indicates that success rate tends to improve as the adversary's token budget increases, and similarly, as the budget increases, the number of steps required to find a match tends to decrease.  Relative to Table~\ref{tab:single-step}, the middle bar ground indicates that halving the budget also serves to reduce the success rate by at least a factor of two.  This indicates that an attacker can reliably obtain stronger control authority over a targeted VLA by increasing its token budget.



\begin{figure}[t]
  \centering

  \begin{minipage}[b]{0.48\textwidth}
    \centering
    \captionof{table}{\textbf{Attacks on real-world images.} We find that our attacks exhibit relatively strong performance when optimized on images drawn from SIMPLER, a simulated environment, and SIMPLER, a real-world environment.}
    \medskip

    \resizebox{\textwidth}{!}{%
      \begin{tabular}{ccc}
        \toprule
        Dimension    & HYDRA (\%) & SIMPLER (\%) \\
        \midrule
        0            & 93.4  & 50.4 \\
        1            & 86.0  & 48.1 \\
        2            & 73.6  & 47.3 \\
        3            & 85.1  & 48.1 \\
        4            & 88.4  & 45.0 \\
        5            & 88.4  & 48.8 \\
        6            & 63.6  & 41.9 \\
        \midrule
        Overall ASR  & 61.2  & 38.0 \\
        \bottomrule
      \end{tabular}
    }
    \label{tab:sim2real-attack-results}
  \end{minipage}%
  \hfill
  \begin{minipage}[b]{0.48\textwidth}
    \centering
    \includegraphics[width=\textwidth]{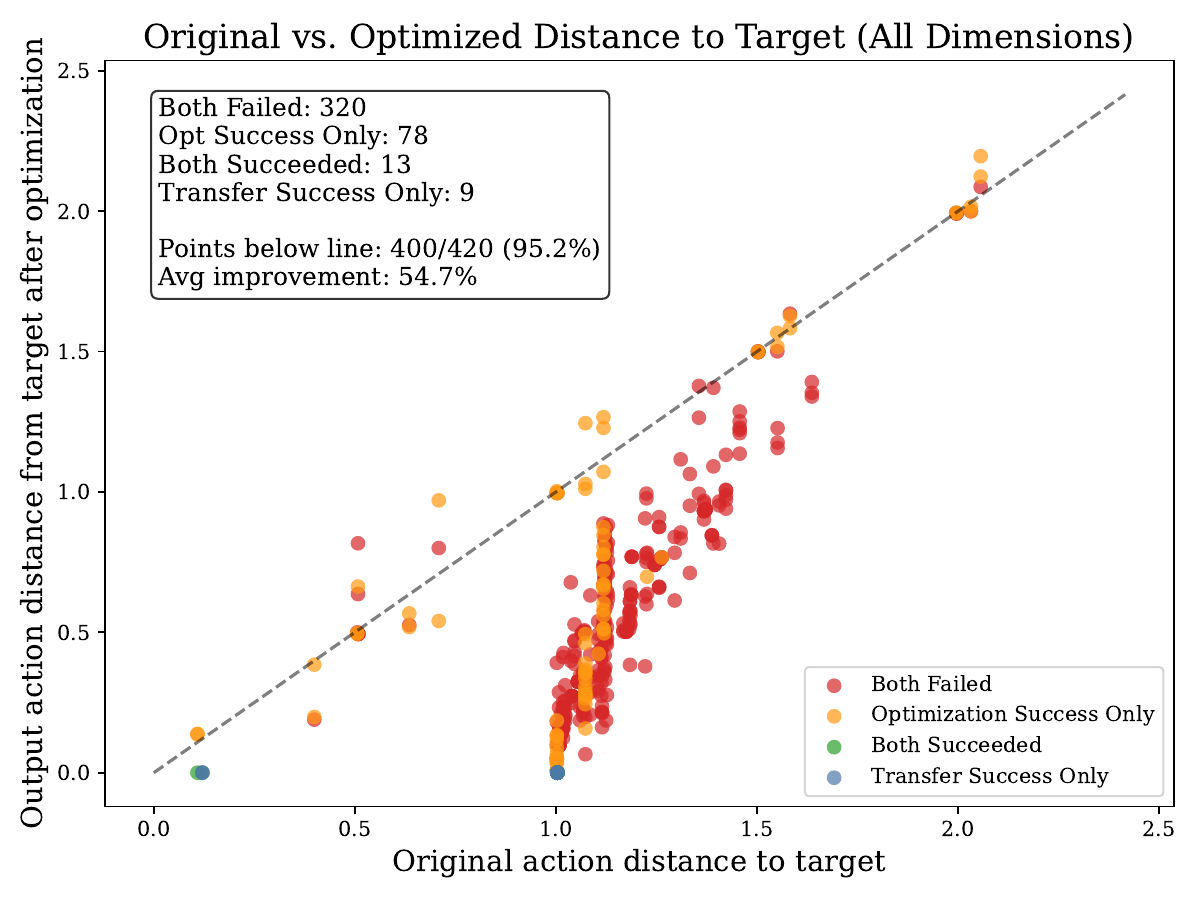}
    \captionof{figure}{\textbf{Ensemble transfer results.} We find that ensemble attacks have a relatively uncorrelated, yet nontrivial effect on transferability.}
    \label{fig:ensemble-transfer}
  \end{minipage}
\end{figure}

\textbf{Attacks are environmentally agnostic.} The so-called ``sim-to-real'' gap in robotics is a phenomenon whereby policies trained in simulated environments struggle to generalize to real-world environments. This challenge extends to VLA policies, which—despite improved generalization from pretraining on large-scale embodied data—still exhibit performance gaps when transferred to real-world settings.  To assess how well our attacks optimized in a simulated environment transfer to real-world settings, we evaluate single-step attacks on two environments from the Open-X-Embodiment~\cite{open_x_embodiment_rt_x_2023} set that OpenVLA was trained on: HYDRA \cite{belkhale2023hydrahybridrobotactions}, a real-world environment, and SIMPLER \cite{li24simpler}, a simulated environment. As shown in Table \ref{tab:sim2real-attack-results}, our attack is successful across both of these environments, indicating that such attacks also yield control authority in more realistic, open-world settings.

\subsection{Persistence attacks}\label{sec:persistence-experiments}

We next consider persistent attacks, for which the goal is to elicit a targeted action over a longer horizon relative to single step attacks.  In this setting, the attacker is given access to $r$ images, where $r\in\{1, 2, 3\}$, which are collected in one of two ways: (1) images are taken from a ``burn-in'' period before the rollout begins, during which the VLA is actuated with randomly generated actions; and (2) images are taken from the first $r$ steps of the rollout.  In both settings, we play the VLA policy for 80 steps after applying the attack.  We use hatches to denote persistence steps corresponding to the $r$ images seen by the attacker, and non-hatched boxes to denote persistence for future, unseen images.  The red dashed line indicates the frequency with which 50 independent, non-attacked rollouts elicit the targeted action.  Our results in Figure~\ref{fig:persistence-overview} indicate that we consistently persist across the seen images, and as we increase the attacker's image budget, generalization to unseen images also tends to increase, particularly on the Libero-Spatial fine-tune.  Moreover, for this model, both variants achieve a nearly 28$\times$ improvement in the number of persistence steps relative to the nominal baseline.
  
\subsection{Transfer attacks}\label{sec:transfer-experiments}

In the setting of transfer attacks, our goals are to (a) evaluate the extent to which attacks optimized for one VLA architecture transfer to other VLA architectures and (b) evaluate whether one can obtain a \textit{universal} attack across $n$ OpenVLA fine-tunes, and assess transfer on the remaining $4-n$ models. 

\paragraph{Architecture transfer.} We optimize single-step attack strings for the OpenVLA base model, and then transfer these strings to three models: TraceVLA \citep{zheng2024tracevlavisualtraceprompting}, CogACT \citep{li2024cogactfoundationalvisionlanguageactionmodel}, and OpenPi0 \citep{githubAllenzrenopenpizero}.  The architectural differences between the chosen models and OpenVLA are discussed in detail in Appendix~\ref{app:architectural-details}. To the best of our knowledge, SIMPLER~\cite{li24simpler} is the only benchmark on which all of these models have been evaluated, and we therefore adopt it for our comparison.

To assess transfer, we compare four different prompting methods: the nominal instruction, a randomly chosen string of tokens from the downstream model's vocabulary, and transferred strings both when the optimization successfully and unsuccessfully resulted in a match on the source model.  In these experiments, we did \emph{not} observe any exact matches across the seven action dimensions.  This is unsurprising, given that textual attacks on standard VLLMs are known to exhibit little, if any, transferability~\cite{schaeffer2024failures}.  We therefore compare the $\ell_2$ distance between the target action and the elicited action (discussed in greater detail in Appendix~\ref{app:architectural-details}). While we find that some attacks resulted in low $\ell_2$ distances, this was more attributable to the fact that these targets happened to be easier to hit, which is evinced by the fact that random prompts tend to do well for these actions.  In other words, optimized instructions tend to do \textit{no better} than random instructions for target action elicitation.

\paragraph{Ensemble transfer.} Due to the computational cost, we only assess ensemble transfer for $n = 2$, meaning we attempt to optimize a universal suffix across two models, and test transfer across the other two. We report our results in Figure~\ref{fig:ensemble-transfer}. We note that transfer has a nontrivial effect on control authority. However, it is uncorrelated with whether the initial optimization over the source models was successful. Further, given that the majority of successful transfers occur at the bottom left of the plot, we observe a relationship between the nominal action's distance from the target and the transfer success rate.  Given the proven ineffectiveness of transfer in multi-modal models, we do not claim to have found conclusive proof of ensemble transfer. However, a perhaps more important takeaway is that we are able to train universal (at $n=2$) attacks that are successful across multiple models. 
\section{Discussion}
\label{sec:defense-discussion}
\textbf{VLA defenses.} Given the connections drawn between chatbot jailbreaking and VLA attacks in \S\ref{sec:vla-attack-algorithms}, a natural question is whether jailbreaking defenses for LLMs and VLLMs can be extended to VLAs.  While several defenses---including those that rely on modified system prompts and in-context demonstrations~\cite{wei2023jailbreak}---are inapplicable given that VLAs do not generally use a system prompt, several earlier defenses can be applied to VLAs.  In particular, we consider the impact of the perplexity filter defense~\cite{jain2023baseline}, which rejects queries if they match a user-defined preplexity threshold, and the smoothing defense \cite{robey2023smoothllm}. In Table~\ref{tab:defense-results}, we record the effectiveness of a text-based and multimodal perplexity filter, and smoothing, for 120 randomly selected one-hot target actions. 

\begin{table}[t]
\centering
\caption{\textbf{Candidate defenses against VLA attacks.} Attack Success Rate (ASR) comparison across models with two modes of perplexity filtering (abbreviated as PF) and smoothing applied.}
\label{tab:defense-results}
\begin{adjustbox}{max width=\columnwidth}
\begin{tabular}{lcccc}
\toprule
Defense & Libero-10 & Libero-Goal & Libero-Object & Libero-Spatial \\
\midrule
No Defense & 63.3 & 100.0 & 96.7 & 100.0 \\
Multimodal PF & 63.3 & 100.0 & 96.7 & 100.0 \\
LLM-Only PF & \textbf{0.0} & \textbf{0.0} & \textbf{0.0} & \textbf{0.0} \\
Smoothing & \textbf{0.0} & \textbf{0.0} & \textbf{0.0} & \textbf{0.0} \\
\bottomrule
\end{tabular}
\end{adjustbox}

\end{table}

As shown, there is a clear difference between the filters when we consider the image embeddings in the loss calculation. The ineffectiveness of the multimodal perplexity filter (where perplexity is calculated over the vision embeddings and the instruction) is due to the dominance of the image inputs over the loss term. We arrive at a similar conclusion as \citet{jain2023baseline}, who find the language-only perplexity filter effective against suffix-style attacks. However, in practice, this defense proves infeasible, primarily because the perplexity threshold depends entirely on the maximum perplexity of instructions seen on a held-out set, which cannot be known beforehand in open-world robotics applications.  Further, we find that smoothing results in a 0\% success rate, but also corrupts the instructions, resulting in a 0\% success rate on non-attacked tasks. It is possible that this will become a more viable defense as models scale in both parameters and capabilities in the future.

\textbf{Safety mechanisms for VLAs.} 
In the field of language modeling, a broad array of techniques are used to align outputs with human intentions, including supervised fine-tuning \citep{bai2022traininghelpfulharmlessassistant}, RLHF \citep{christiano2023deepreinforcementlearninghuman, leike2018scalableagentalignmentreward}, and adversarial training \citep{mazeika2024harmbenchstandardizedevaluationframework}.  However, due to a mix of limited capability sets and decreased attention relative to chatbots, the field of AI-enabled robotics has not yet seen the proposal of analogous notions of alignment.  Initial efforts, such the GRAPE robotic alignment algorithm~\cite{zhang2024grape}, align VLAs at the trajectory-level toward improving generalization and task completion.  In a similar spirit, several works design RL-style approaches to improve robot safety~\cite{abbas2024safetydrivendeepreinforcementlearning,zhang2025safevla}.  However, these works do not consider adversarial attacks on VLAs. This points to the need for notions of VLA refusal when attempts to subvert control are detected.  Moreover, future defenses could incorporate tools from classical control (e.g., control barrier functions or formal methods), which have recently shown effectiveness against attacks on robotic planners~\cite{ravichandran2025safety}.

\section{Conclusion}

VLAs are gaining momentum in the field of robotics due to their ability to fuse the textual and visual understanding of VLMs with the low-level actuation.  In this paper, we attempt to anticipate future threat models that may impact robotic foundation models as they are deployed commercially.  In particular, we present the first study of adversarial attacks on low-level VLA actuators, showing that by optimizing instructions we can obtain complete control authority over a target VLA. These results underline the necessity for new forms of defenses that are reflective of the unique output format VLAs pose, as these systems become more powerful and widely used in society.  

\paragraph{Limitations and future work.} 
\label{limitations}
We recognize that our attack may be difficult to employ in practice, due to the white-box nature and relative cost of the GCG algorithm. Further, while our attack is designed to work on any autoregressive VLA, diffusion-based models are also very prevalent throughout the field. Extending attack frameworks to black-box scenarios and diffusion-based models will be a critical step in the pursuit of fully assessing the risks these models pose.

\newpage
\bibliographystyle{unsrtnat}
\bibliography{bibliography}
\newpage
\appendix

\section{On the consequences of fine-tuning}
\label{appendix:a}

\begin{figure}[!ht]
    \centering
    \includegraphics[width=0.9\linewidth]{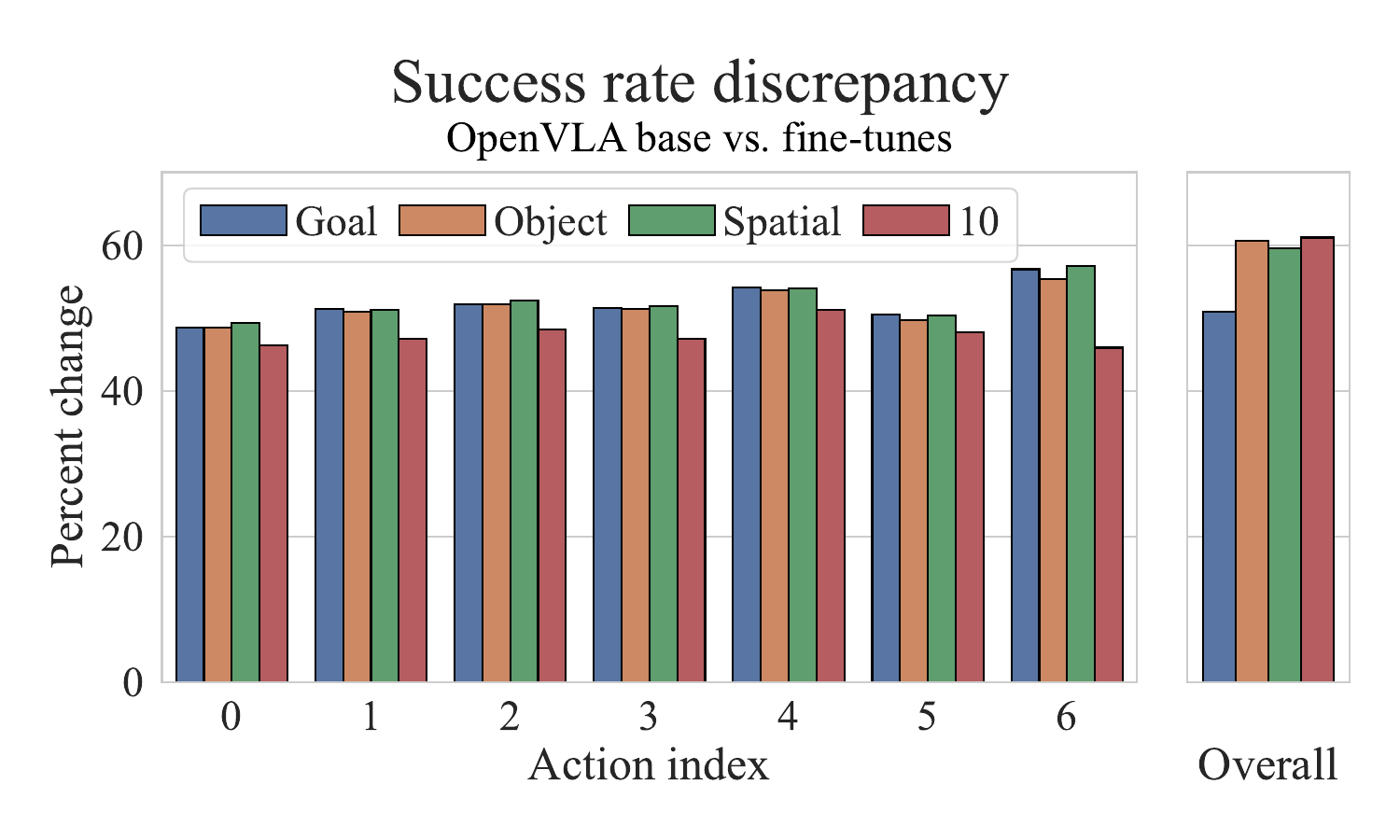}
    \vspace{-1.5em}
    \caption{\textbf{Attacking fine-tuned vs.\ base VLAs.} Each bar shows the percent change in the single step access rate of each OpenVLA fine-tune relative to analogous attacks on the OpenVLA base model.  Both the overall and per-dimension success rates show a similar trend: single step attack effectiveness increases by 40-60\% for each fine-tune relative to the base model.}
    \label{fig:discrepancy}
\end{figure}

\begin{figure}[H]
    \centering
    \includegraphics[width=\linewidth]{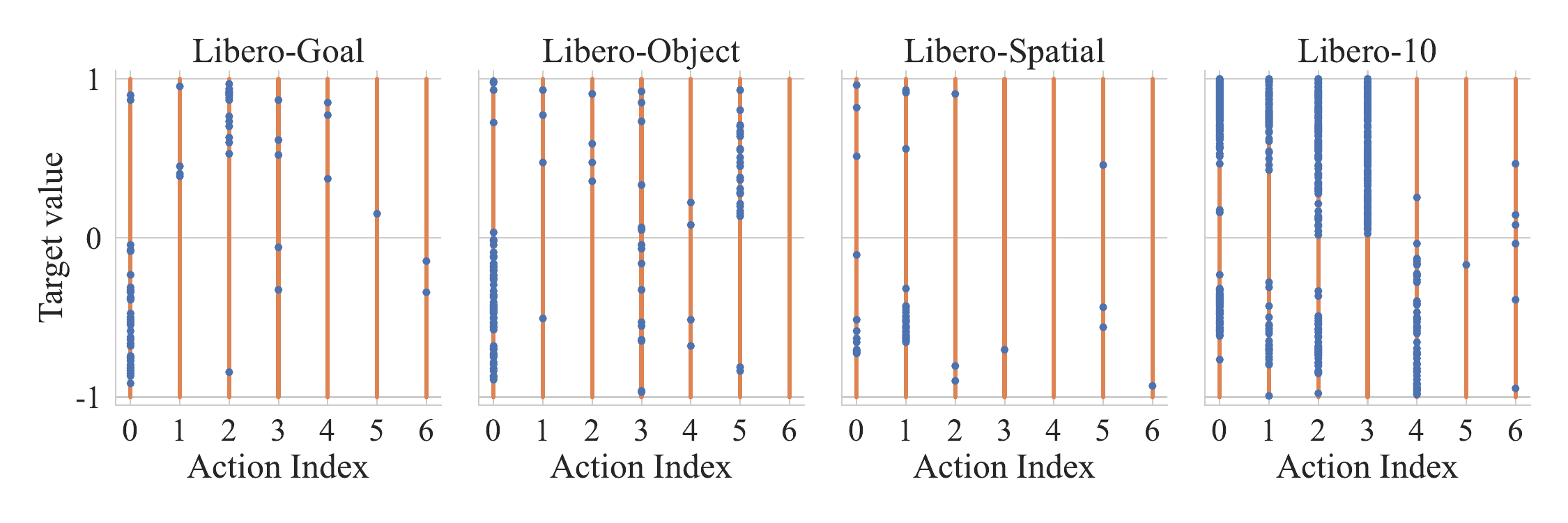}
    \caption{\textbf{Visualizing single step success rates.} As we use one-hot target actions in the evaluation of single step attacks in \S\ref{sec:single-step-experiments}, we can visualize the locations at which the attack fail. In particular, the $x$-axis of these plots shows the action dimensions, and the $y$-axis shows the value of the of the one-hot component of the target.  The blue dots denote the locations at which the single step attacks failed.  We observe significant clustering of failures, particularly for the Libero-10 fine-tune, which had the lowest overall success rate.}
    \label{fig:single-step-grid}
\end{figure}

 While the paradigm of fine-tuning VLAs on downstream tasks has gained traction of late~\cite{kim24openvla,stachowiczlifelong}, earlier works at the intersection of VLAs and robotics tended to focus on employing generalist, non-fine-tuned policies~\cite{open_x_embodiment_rt_x_2023,driess2023palme}.  Thus, a natural question is whether task-specific fine-tuning has an impact on the effectiveness of attacks on VLAs.  In Figure~\ref{fig:discrepancy}, we find that the OpenVLA base model is significantly more resistant to adversarial attacks relative to the four fine-tunes in Table~\ref{tab:single-step}. More specifically, across a uniform gridding of the action space of the OpenVLA-Base model, we ran the single step attack on 120 different target actions, recording our overall success rate of only 38\% (see Table~\ref{tab:sim2real-attack-results}). In Figure~\ref{fig:discrepancy}, we plot the percent increase in the effectiveness of our single step attack on each of the four OpenVLA fine-tunes relative to the base model. We find that on average, attack effectiveness increases by nearly 40-60\% for the fine-tuned models relative to the base model.  This could indicate that fine-tuning in in some sense concentrates the actions in a smaller region (after normalization), making attacks easier to carry out.  This point is corroborated by the evidence in Figure~\ref{fig:single-step-grid}, wherein we visualize the action spaces of the four fine-tuned models; we observe noticeable clustering of the locations of single step attacks, particularly for the lowest performing Libero-10 model.
\section{Continued discussion of architectural transfer}
\label{appendix:b}

\begin{figure}[!ht]
    \centering
    \includegraphics[width=0.9\linewidth]{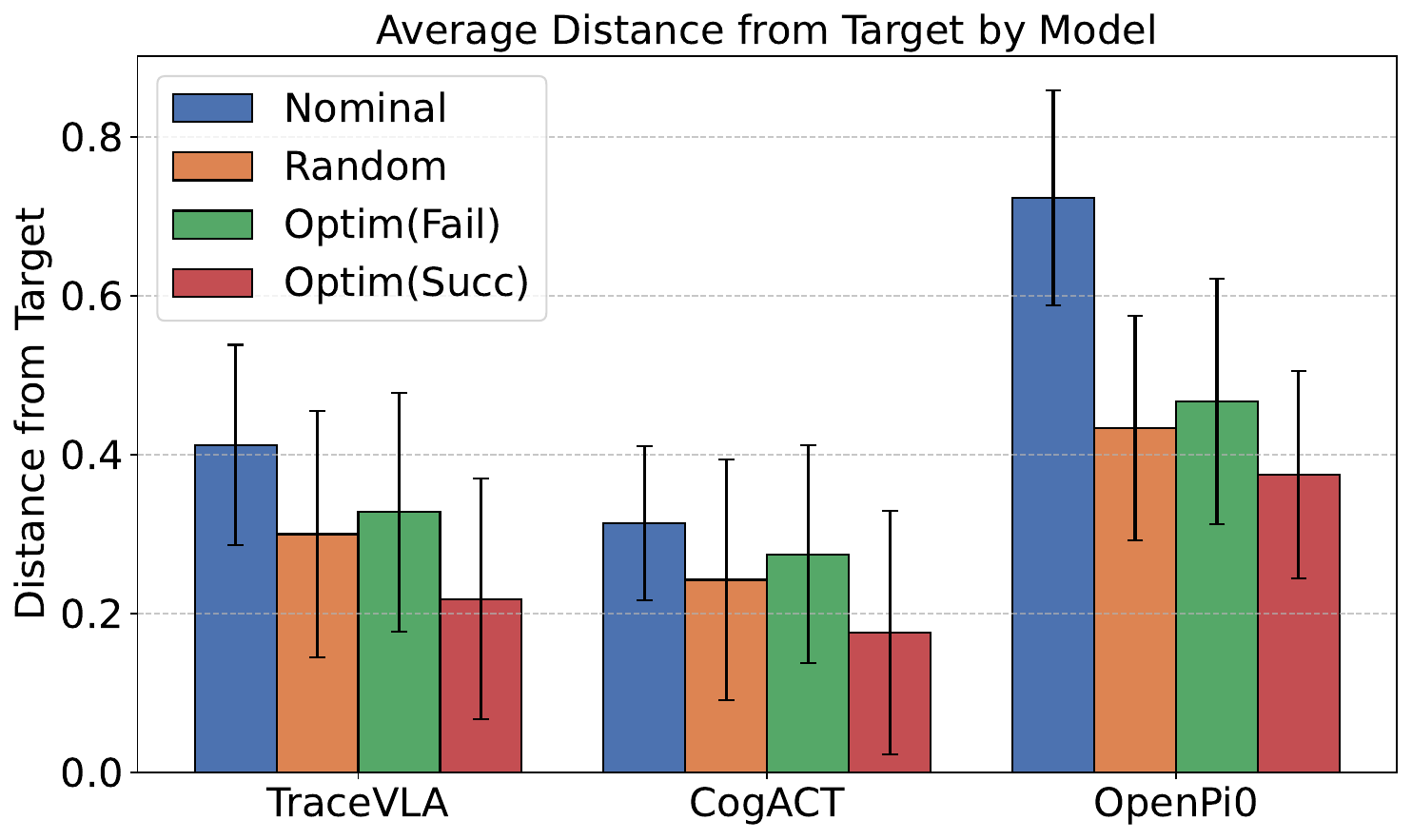} 
    \vspace{-1em}
    \caption{\textbf{Transfer attacks.} By optimizing an instruction for OpenVLA-Base, and then attempt transfer to each of TraceVLA, CogACT, and OpenPi0, we evaluate the effectiveness of transferable attacks. Our results show that there is little to no transfer between OpenVLA-Base and other models.}
    \label{fig:transfer}
\end{figure}

\subsection{Architectural details}
\label{app:architectural-details}
We begin by discussing the differences between OpenVLA and the models chosen for our testing of transfer. For our motivations behind choosing models with different architectures, please refer to appendix \ref{appendix:a}

Of the three models chosen to test transfer against, \textbf{TraceVLA} \citep{zheng2024tracevlavisualtraceprompting} is the closest architectural match. It is a direct finetune of OpenVLA, however at each step, the model is given two images: the original image, as well as a version of the original image overlaid with a visual trace of active point trajectories. Therefore, at each step $t$, the model requires two images instead of one, therefore drastically shrinking the percentage of the embedding space occupied by the text embeddings. \textbf{CogACT} \citep{li2024cogactfoundationalvisionlanguageactionmodel} introduces a diffusion action module onto the end of the language model outputs in order to predict a sequence of actions. Critically, the diffusion-based architecture makes text-based optimizations difficult due to the more continuous nature of the outputs. Finally, \textbf{OpenPi0} \citep{githubAllenzrenopenpizero}, an open-source implementation of $\pi_0$ \citep{black2024pi0visionlanguageactionflowmodel}, implements an \textit{action expert} on top of the language model, which utilizes both a diffusion-based process and a separate set of weights for improved performance. All three models output actions in seven dimensions, identically to OpenVLA. 

\subsection{Random vs. optimized instruction action distributions}

In this subsection, we further discuss our conclusion that transfer between models of different architectures has not occurred during our experiments. All experiments were run offline using an image taken from the SIMPLER environment for the "pick coke can" objective. For a given rollout, the image used for both GCG optimization and our tests for transfer was taken from timestep 10, which is the first image after the burn-in period.

In the figures below: 
\begin{itemize}
    \item "Nominal" refers to the nominal action that is predicted by each of the models when provided the "nominal" instruction, which is "pick coke can". 
    \vspace{-0.5em} 
    \item "Optimized" refers to the action that is predicted by each of the models when provided the "optimized" string that is a result of running GCG on OpenVLA, under the nominal "pick coke can" environment. We use "optimized instruction" to refer to the instruction itself, and "optimized action" to refer to the resulting action.
    \vspace{-0.5em}
    \item "Random" refers to the action that is predicted by each of the models when provided with a "random" string sampled from each model's vocabulary as the instruction. 
    \vspace{-0.5em} 
    \item "Target" refers to the action that we optimized fore when running the GCG optimization algorithm on OpenVLA. This was also the target action that we compared each of the three models' actions to when computing $\ell_2$ distance.  
\end{itemize}

\begin{figure}[H]
    \centering
    \includegraphics[width=\linewidth]{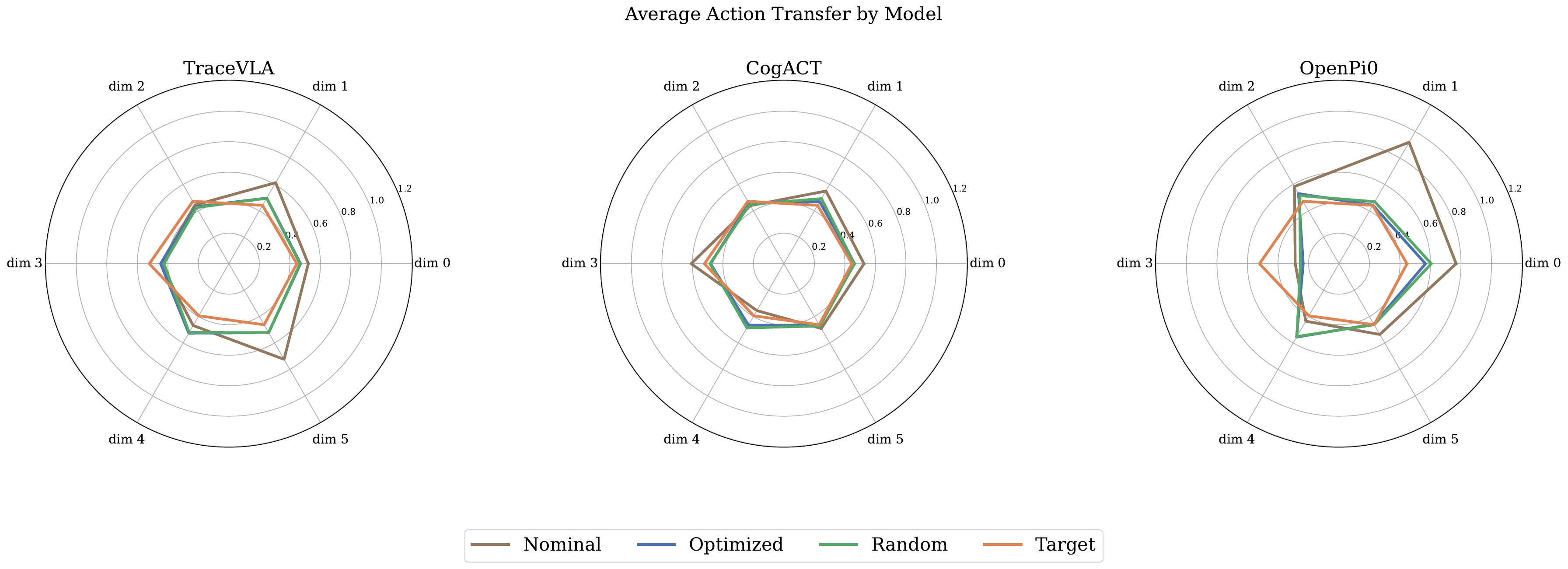}
    \caption{\textbf{Average action for samples where GCG \textit{did not} converge}}
    \label{fig:nonconvergence}
\end{figure}

\begin{figure}[H]
    \centering
    \includegraphics[width=\linewidth]{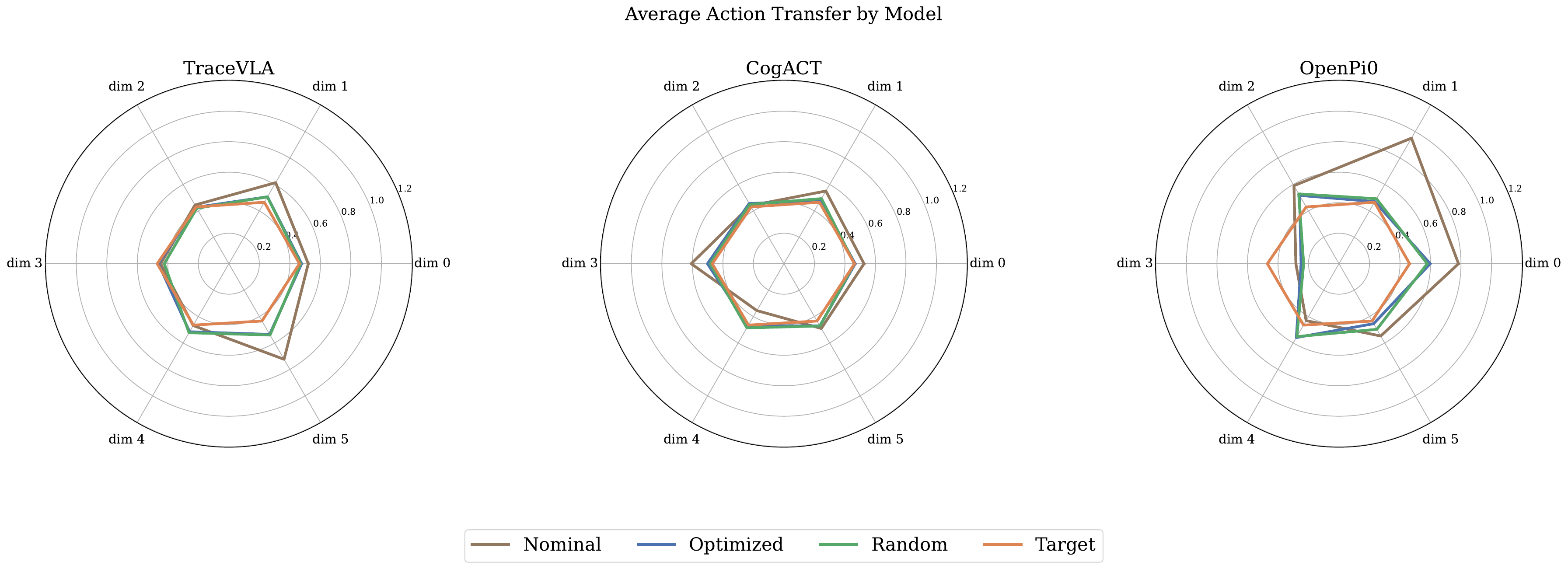}
    \caption{\textbf{Average action for samples where GCG \textit{did} converge}}
    \label{fig:convergence}
\end{figure}

Figure \ref{fig:nonconvergence} presents the average action across the first six dimensions (disregarding the gripper, which skews the distributions of the actions due to its relatively large magnitude) for each model, comparing the nominal, unsuccessfully optimized, and random instructions to the target action. Figure \ref{fig:convergence} presents the average action across the first six dimensions, except for all runs where GCG converged to obtain an instruction that is optimized for the target action. The plots have been normalized such that there are no negative actions, despite the fact that the action space lies in $[-1, 1]$.

Figure \ref{fig:transfer} alone appears to demonstrate that the optimized instructions play an integral role in the success of transfer. However, in examining these figures, a different conclusion can be made. First, across both successful and unsuccessful optimizations, the average action elicited from the optimized instruction is incredibly similar to the average action elicited from random instructions. This suggests that the optimized strings are being interpreted no differently than random strings of tokens, and result in similar action outputs. Secondly, the average target action, for samples where GCG did converge, is far closer to the average actions for the optimized and random instructions than it is for samples were GCG did not converge. This is best evidenced in the TraceVLA plots, where we can clearly see the degree to which the target action follows a similar distribution to the optimized and random actions in the successful case, compared to the unsuccessful one. Further, we notice that for samples where GCG converged, the average target action is much more uniform compared to the samples where GCG did not converge. These facts yield the following conclusions: 

\begin{itemize}
    \item \textbf{Whether or not a GCG run converged is correlated with the uniformity of the target action about the $\vec{0}$ vector.} For the OpenVLA base model that was not finetuned on downstream LIBERO tasks, we found that one of the most accessible target actions was $\vec{0}$, and as such, the randomly selected actions that we used to test transfer were most likely to converge the closer they were to that vector.
    \item \textbf{Optimized instructions are seen no differently to random instructions in the vocabularies of the models that we tested.} This is demonstrated by the almost identical average actions for optimized instructions and random instructions across both figure \ref{fig:convergence} and figure \ref{fig:nonconvergence}.
    \item \textbf{The seemingly positive relationship between GCG convergence and ability for the optimized instructions to transfer is due to the randomly selected target action's proximity to the distribution of random and optimized actions.} It is not reflective of successful transfer, despite at first glance appearing to be. 
\end{itemize}
\section{Ablations}
\label{appendix:b}

\subsection{Action elicitation}
We notice that in figure \ref{fig:single-step-grid}, target actions where the optimization failed are likely to be grouped together. Most notably, for Libero-10, the worst-performing of the four finetunes, we notice that there are large sections of the action space that are difficult to attain. In part, this can be attributed to the normalization factor: across the first four action indices, the q01 values are $-0.6348214149475098$, $-0.7741071581840515$, $-0.7633928656578064$, and $-0.09749999642372131$ respectively. Best illustrated at index 3, since the q01 value is so close to 0, almost all 128 actions between 0 and -1 get normalized to this value. As a result, we are effectively repeating the same optimization trial for a singular value repeatedly until we reach $-0.09749999642372131$, an action which is clearly difficult to achieve in the action space. This is represented in the figure, as nearly all of those trials are failures. Similarly for dimensions 0 and 1, the majority of failures on that half of the action space occur before we reach the q01 value. This is not always the case, however. Note that, aside from the fourth dimension of Libero-10 and the 0th dimension of Libero-Object, we are able to achieve maximum values across all other dimensions of all other models. 

\begin{table}[!ht]
    \centering
    \caption{\textbf{Suffix Trials.} We report the per-dimension ASR and overall ASR for each of the four models across 12 different actions, the $-1$ and $1$ one-hot vectors across the first six dimensoins (excluding the gripper), when utilizing the optimized instruction as a suffix for the nominal instruction.}
    \label{tab:suffix-trials}
    \vspace{2mm}
    \begin{adjustbox}{max width=\textwidth}
    \begin{tabular}{cccccccccc} \toprule
         \multirow{2}{*}{Model} & \multicolumn{6}{c}{Per-dimension success rate} & \multirow{2}{*}{\makecell{Overall \\ success rate}} & \multicolumn{2}{c}{Avg.\ computation per success} \\ \cmidrule(lr){2-7} \cmidrule(lr){9-10}
         & 0 & 1 & 2 & 3 & 4 & 5 & & Optim.\ steps & Time (sec.) \\ \midrule
         Libero-Goal & 100.0 & 100.0 & 91.7 & 100.0 & 100.0 & 100.0 & 91.7 & 170.2 & 942.7 \\
         \midrule
         Libero-Object & 100.0 & 91.7 & 91.7 & 91.7 & 75.0 & 91.7 & 50.0 & 184.7 & 1242.5 \\
         \midrule
         Libero-Spatial & 91.7 & 83.3 & 91.7 & 100.0 & 91.7 & 91.7 & 75.0 & 121.0 & 678.0 \\
         \midrule
         Libero-10 & 83.3 & 75.0 & 91.7 & 83.3 & 91.7 & 75.0 & 41.7 & 238.2 & 1358.3 \\
         \bottomrule
    \end{tabular}
    \end{adjustbox}
\end{table}

\subsection{Suffix trials}
Table \ref{tab:suffix-trials} demonstrates the success of our method when being utilized akin to standard GCG, where we optimize a suffix to be placed at the end of an instruction. The nominal instructions for the chosen tasks are as follows:

\begin{itemize}
    \item \textbf{Libero-Goal:} ``push the plate to the front of the stove''
    \item \textbf{Libero-Object:} ``pick up the alphabet soup and place it in the basket''
    \item \textbf{Libero-Spatial:} ``pick up the black bowl between the plate and the ramekin and place it on the plate''
    \item \textbf{Libero-10:} ``pick up the book and place it in the back compartment of the caddy''
\end{itemize}

For the first 6 action dimensions, we perform GCG optimization on the maximum ($1$) and minimum ($-1$) one-hot vectors, as we find these to be the most difficult actions to achieve within the action space. Our results show with certainty that our method works not only as an optimized instruction, but also as an optimized suffix, given that additional trials on easier actions will result in a much higher true ASR, provided an increase in budget. Finally, in figure \ref{fig:token-ablation}, we show that as an attacker's token budget increases, so do overall success rates, and that increasing a budget also leads to fewer optimization steps being necessary for convergence. 

\newpage

\end{document}